%% file: paper.tex
\documentclass[letterpaper, 10 pt, conference]{ieeeconf}  

\IEEEoverridecommandlockouts                              

\usepackage{graphics} 
\usepackage{epsfig} 
\usepackage{times} 
\usepackage{amsmath} 
\usepackage{amssymb}  
\usepackage{bibentry}
\usepackage{tikz}
\usepackage{microtype}
\usepackage{svg}
\usepackage{algorithm}
\usepackage{algorithmic}
\usepackage{multirow}
\usepackage{colortbl}
\usepackage{booktabs}
\usepackage{balance}
\usepackage[hidelinks]{hyperref}
\usepackage{eso-pic} 

\title{\LARGE \bf
Unified Multi-Rate Model Predictive Control for a Jet-Powered Humanoid Robot
}

\author{Davide Gorbani$^{1,2}$, Giuseppe L'Erario$^{1}$,   Hosameldin Awadalla Omer Mohamed$^{1}$, Daniele Pucci$^{1,2}$
\thanks{$^{1}$Artificial and Mechanical Intelligence, Istituto Italiano di Tecnologia, Genoa, Italy {\tt\small firstname.surname@iit.it}}%
\thanks{$^{2}$School of Computer Science, University of Manchester, Manchester, UK        }}

\include{variablesNamingShortcuts}

\begin{document}

\AddToShipoutPictureBG*{%
  \AtPageUpperLeft{%
    \setlength{\unitlength}{1mm}%
    \put(0,-12){\makebox(\paperwidth,0)[c]{\parbox{0.8\textwidth}{\centering\textcolor{gray}{\large This paper has been accepted for publication at the 2025 IEEE-RAS 24th International Conference on Humanoid Robots (Humanoids), Seoul, 2025.  ©IEEE}}}}
  }
}

\maketitle
\thispagestyle{empty}
\pagestyle{empty}


\input{sections/abstract}
\input{sections/introduction}

\input{sections/background}

\input{sections/method}

\input{sections/results}

\input{sections/conclusions}
\input{sections/appendix}




\balance
\bibliographystyle{unsrt}
\bibliography{paper}






\end{document}

%% file: variablesNamingShortcuts.tex
\newcommand{\frameW}{\mathcal{J}}

\newcommand{\frameCoM}{\mathcal{G}}

\newcommand{\frameB}{\mathcal{B}}

\newcommand{\frameCent}{\mathcal{\frameCoM[\frameW]}}

\newcommand{\frameMixed}{\mathcal{\frameCoM[\frameB]}}



%% file: sections/abstract.tex
\begin{abstract}

We propose a novel Model Predictive Control (MPC) framework for a jet-powered flying humanoid robot. The controller is based on a linearised centroidal momentum model to represent the flight dynamics, augmented with a second-order nonlinear model to explicitly account for the slow and nonlinear dynamics of jet propulsion. A key contribution is the introduction of a multi-rate MPC formulation that handles the different actuation rates of the robot’s joints and jet engines while embedding the jet dynamics directly into the predictive model. We validated the framework using the jet-powered humanoid robot iRonCub, performing simulations in Mujoco; the simulation results demonstrate the robot’s ability to recover from external disturbances and perform stable, non-abrupt flight manoeuvres, validating the effectiveness of the proposed approach.\looseness=-1

\end{abstract}

%% file: sections/introduction.tex
\section{INTRODUCTION}

Humanoid robots have traditionally been designed for terrestrial locomotion, mimicking human-like walking and manipulation. However, the advent of flying humanoid robots introduces a fascinating new paradigm — robots that not only resemble human morphology but also possess the ability to fly. This hybrid capability opens up a range of novel applications in disaster response, search and rescue, and exploration, where both agile ground interaction and aerial mobility are essential.\looseness=-1

In literature, there are different examples of aerial robots equipped with manipulation or locomotion capabilities \cite{zhou2024current}; for instance, aerial manipulators consist of a flying platform equipped with a robotic arm or gripper \cite{khamseh2018aerial, ollero2021past}.\looseness=-1
While aerial manipulators are a common example of flying multibody robots, they are not the only type. Other examples include hexapod–quadrotor hybrids \cite{Pitonyak2017}, ground‐and‐air capable quadrotors \cite{Kalantari2013}, insect‐inspired biobots capable of both flying and walking \cite{Bozkurt2009}, shape‐shifting robots that adapt their morphology for different locomotion modes \cite{Daler2015, Daler2013}, and humanoid robots equipped with thrusters to support bipedal motion \cite{Kim2021, Huang2017}. An effort to unify manipulation, aerial mobility, and terrestrial locomotion on a single platform is being carried out by the iRonCub project, which aims to enable the humanoid robot iCub 3 \cite{dafarra2024icub3} to fly by adding four jet engines on its arms and shoulders.\looseness=-1

\begin{figure}[t]
    \centering
    \includegraphics[width=1.0\linewidth]{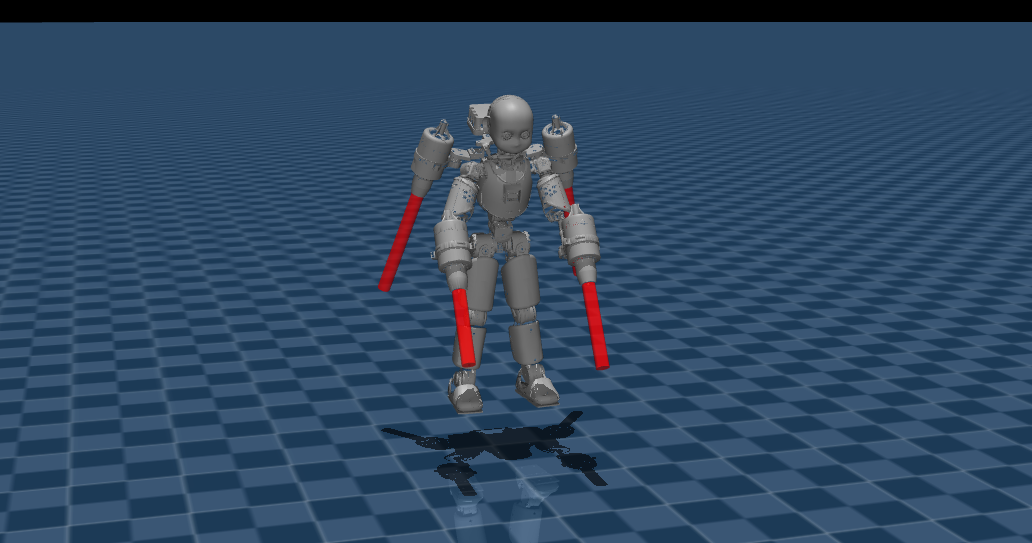}
    \caption{iRonCub robot in Mujoco simulator; the thrust forces provided by the jets are visualised as red cylinders.}
    \label{fig:enter-label}
    \vspace{-0.32cm}
\end{figure}

Controllers for aerial vehicles have been extensively studied, with strategies ranging from classical PID control to advanced nonlinear and optimal control techniques and data-driven techniques \cite{zuo2022unmanned}, \cite{besnard2012quadrotor}, \cite{torrente2021data}. Other examples of control techniques used for aerial systems are controllers that exploit the \textit{thrust-vectoring} for quadrotors with tilting propellers~\cite{kumar2020quaternion}.\looseness=-1

The above-mentioned controllers are not designed for articulated robots but for single-body aerial systems, such as quadcopters; previous research on jet-powered humanoid control has primarily focused on momentum-based strategies. \cite{pucci2017momentum} proposes a framework for asymptotically stabilising the centroidal momentum of the iCub robot, employing a vectored-thrust approach within a hierarchical control structure. Building upon this, \cite{nava2018flightController} introduced a task-based controller capable of tracking full position and attitude trajectories, addressing prior limitations in angular momentum handling and improving flight orientation control. 
\cite{taliani2025nlmpc}~proposes a nonlinear MPC to stabilise both legged and aerial phases of a trajectory performed by the robot.

However, all of these controllers neglected the dynamics of the jet engines, whose behaviour is highly nonlinear, as highlighted in \cite{lerario2020modeling} and \cite{momin2022nonlinear}, where the jets were modelled as second-order nonlinear systems. To incorporate jet dynamics in flight control strategies, \cite{lerario2022trajectoryopt} proposed an approach for trajectory generation; nevertheless, this method remains limited by its offline nature and reliance on a whole-body controller for stabilisation, rendering it unsuitable for real-time control.\looseness=-1

A more recent solution was proposed in \cite{lerario2024learning}, where a Reinforcement Learning framework was developed to enable the robot to walk and fly, explicitly accounting for jet dynamics. However, this framework requires precise torque-level control, which is not available on all robotic platforms, and it neglects the limitations posed by the low bandwidth actuation of the jets.\looseness=-1

To address the challenge of designing an online flight controller that accounts for the complex dynamics of jet propulsion, we propose a Model Predictive Control (MPC) framework that integrates the centroidal momentum dynamics of the robot \cite{orin2013centroidal} with the second-order nonlinear jet model introduced in \cite{lerario2020modeling}. To ensure compatibility with high-frequency online requirements, both the centroidal and jet dynamics are linearised, resulting in a Linear Parameter-Varying (LPV) MPC formulation.\looseness=-1

Moreover, to manage the differing control frequencies of the robot's joints and jet engines, we introduce a strategy that allows the MPC to compute control inputs at distinct rates for each actuator type. While previous frameworks have addressed multi-rate systems using hierarchical or cascaded control architectures \cite{rosolia2020multi}, \cite{farina2018hierarchical}, or have focused on systems with mismatched sampling and actuation periods \cite{ling2004model}, our approach proposes a unified MPC framework that natively generates control commands at multiple frequencies without relying on hierarchical decomposition.\looseness=-1

The remainder of the paper is organised as follows: Section~\ref{sec:background} introduces the notation and reminds the \textit{floating base} formalism; Section \ref{sec:method} explains the mathematical formulation of the proposed MPC; Section \ref{sec:results} shows the results achieved in simulation on the flying humanoid robot iRonCub; Section \ref{sec:conclusion} concludes the paper with limitations and future directions.\looseness=-1

%% file: sections/background.tex
\section{BACKGROUND}
\label{sec:background}

\subsection{Notation}

\begin{itemize}
    \item $\frameW$ denotes the world frame;
    \item $\frameB$ denotes the body frame;
    \item $\frameMixed$ denotes a frame with the origin in the centre of mass and the orientation of the body frame;
    \item $\frameCent$ denotes a frame with the origin in the centre of mass and the orientation of the inertial frame;
    \item $^{\mathcal{A}}R_{\mathcal{B}}$ is the rotation matrix from frame $\mathcal{A}$ to frame $\mathcal{B}$;
    \item $^{\mathcal{A}}p_{\mathcal{B},\mathcal{C}} \in \mathbb{R}^{3}$ denotes the vector  connecting the origin of the frame $\mathcal{B}$ to the origin of frame $\mathcal{C}$ expressed in frame $\mathcal{A}$;
    \item $^{\mathcal{A}}\omega_{\mathcal{B}}$ is the angular velocity of frame $\mathcal{B}$ respect to the inertial frame $\frameW$ expressed in frame $\mathcal{A}$;
    \item $S(x) \in \mathbb{R}^{3 \times 3}$ is the skew-symmetric matrix such that $S(x)y = x \times y$, where $\times$ is the cross product operator in $\mathbb{R}^{3}$;
\end{itemize}

\subsection{Floating-base model}
A flying humanoid robot can be modelled as a multi-body system composed of $n+1$ rigid bodies, called links, connected by $n$ joints with one degree of freedom. Following the \textit{floating base} formalism, it is possible to define the configuration space $\mathbb{Q}\in\mathbb{R}^{3}\times SO(3) \times \mathbb{R}^{n}$; an element of the space is composed of the tuple $q = ({}^{\frameW}p_{\frameW, \frameB},{}^{\frameW}R_{\frameB},s)$. The velocity belongs to the space $\mathbb{V}\in \mathbb{R}^{3}\times \mathbb{R}^{3}\times \mathbb{R}^{n}$ and an element of the space $\mathbb{V}$ is $\nu = ({}^{\frameW}\dot{p}_{\frameW, \frameB}, {}^{\frameW}\omega_{\frameB}, \Dot{s})$. By applying the Euler-Poincaré formalism, the equation of motion for a multi-body system subject to \textit{m} external forces becomes \cite{traversaro2017modelling}:
\begin{equation}
    M(q)\dot{\nu} + C(q, \nu)\nu + G(q) = 
    \begin{bmatrix}
    0_{6 \times 1} \\
    \tau
    \end{bmatrix}
    + \sum_{k=1}^{m} J_k^\top F_k
    \label{eq:floatingBase},
\end{equation}

where the matrices $M, C \in \mathbb{R}^{(n+6)\times(n+6)}$ are the mass and the Coriolis matrix, the term $G \in \mathbb{R}^{n+6}$ is the gravity vector, the term $\tau \in \mathbb{R}^{n}$ are the internal actuation torques; $F_{k}$ is the is the \textit{k}th of \textit{m} external forces applied on a frame $\mathcal{C}_{k}$ attached to the robot and $J_{k}$ is the Jacobian that maps the velocity of the robot $\nu$ into the velocity of the frame ${}^{\frameW}\Dot{p}_{\mathcal{C}_{k}}$ of the origin of $\mathcal{C}_{k}$.

%% file: sections/method.tex
\section{METHOD}
\label{sec:method}

\subsection{Centroidal momentum equations}
\label{subsec:centMomEq}
To formulate the MPC, we chose to use the centroidal momentum equation rather than the equation of motion in \eqref{eq:floatingBase}. This approach models the robot’s dynamics with fewer variables, leading to a smaller optimisation problem that demands less computational effort and time. As a result, the proposed controller becomes more suitable for real-time implementation on the robot. The \textit{centroidal momentum} equation is \cite{orin2013centroidal}:
\begin{equation}
    ^{\frameCent} {h} =
    \begin{bmatrix}
        ^{\frameCent} {h}^{p} \\
        ^{\frameCent} {h}^{w}
    \end{bmatrix}
    = 
    \begin{bmatrix}
        m {^{\frameW}v_{\frameW,\frameCoM}} \\
        ^{\frameCent} \mathbb{I} {^{\frameW}\omega_{o}}
    \end{bmatrix} \in \mathbb{R}^{6},
    \label{eq:centroidalMom}
\end{equation}
where $m$ is the mass of the robot, $^{\frameCent} {h}^{p}$ is the linear momentum, $^{\frameCent} {h}^{w}$ is the angular momentum, $^{\frameCent} \mathbb{I}(q) \in \mathbb{R}^{3\times3}$ is the robot total inertia expressed in the frame $\frameCent$ and $^{\frameW}\omega_{o}$ is the so called \textit{locked} (or average) angular velocity \cite{orin2013centroidal}. As proposed in \cite{nava2018flightController}, premultiplying \eqref{eq:centroidalMom} by the rotation matrix ${^{\frameB}R_{\frameW}}$, we obtain a change of coordinate in $\frameMixed$: 

\begin{equation}
    ^{\frameMixed} {h} =
    \begin{bmatrix}
        {^{\frameB}R_{\frameW}} {^{\frameCent} {h}^{p}} \\
        {^{\frameB}R_{\frameW}} {^{\frameCent} {h}^{w}}
    \end{bmatrix}
    = 
    \begin{bmatrix}
        m {^{\frameB}v_{\frameW,\frameCoM}} \\
        ^{\frameMixed} \mathbb{I} {^{\frameB}\omega_{o}}
    \end{bmatrix},
    \label{eq:centroidalMomBody}
\end{equation}
where $^{\frameMixed} \mathbb{I} = {^{\frameB}R_{\frameW}} {^{\frameCent} \mathbb{I}} {^{\frameW}R_{\frameB}}$ is the total inertia in new coordinates.

The time derivative of \eqref{eq:centroidalMomBody} is:
\begin{equation}
    ^{\frameMixed} \dot{h} =
    \begin{bmatrix}
        {^{\frameB}\dot{R}_{\frameW}} {^{\frameCent} {h}^{p}} + {^{\frameB}R_{\frameW}} {^{\frameCent} \dot{h}^{p}} \\
        {^{\frameB}\dot{R}_{\frameW}} {^{\frameCent} {h}^{w}} + {^{\frameB}R_{\frameW}} {^{\frameCent} \dot{h}^{w}}
    \end{bmatrix},
    \label{eq:centroidalMomBodyDerivative}
\end{equation}
The terms ${^{\frameB}R_{\frameW}} {^{\frameCent}\dot{h}^{p}}$ and ${^{\frameB}R_{\frameW}} {^{\frameCent}\dot{h}^{w}}$ are defined as:
\begin{align}
    {^{\frameB}R_{\frameW}} {^{\frameCent}\dot{h}^{p}} =& A_{lin}(s)T + mg{^{\frameB}R_{\frameW}e_{3}} \label{eq:eq1}, \\
    {^{\frameB}R_{\frameW}} {^{\frameCent}\dot{h}^{w}} =& A_{ang}(s)T ,\label{eq:eq2}
\end{align}
where $s$ represents the joint positions and ${T}$ is the vector collecting the thrust forces ${T=[T_{1},\dots,T_{n_{j}}]}$ of the $n_{j}$ jet turbines, the matrix ${A}$ is defined as:
\begin{align}
    {A}_{lin}({s}) &= \begin{bmatrix} ^{\frameB} R_{J_1}e_3  \quad  \dots \quad  ^{\frameB} R_{J_4}e_3 \end{bmatrix}, \\
    A_{ang}(s) & = 
    \begin{bmatrix}
    S(^{\frameB}p_{J_1, \frameCoM})^{\frameB}R_{J_1}e_3 \dots  S(^{\frameB}p_{J_4, \frameCoM})^{\frameB}R_{J_4}e_3,
    \end{bmatrix}
\end{align}
and the vector ${e}_3$ is:
\begin{equation*}
    {e}_3 = 
    \begin{bmatrix}
        0 & 0 & 1
    \end{bmatrix} ^{\top}.
\end{equation*}
Exploiting the definition of the derivative of the rotation matrix $^{\frameB}\dot{R}_{\frameW}=-S(^{\frameB}\omega_{\frameB}){^{\frameB}{R}_{\frameW}}$ and combining  \eqref{eq:centroidalMomBodyDerivative}, \eqref{eq:eq1}, and \eqref{eq:eq2}, we get:
\begin{equation}
    ^{\frameMixed} \dot{h} =
    \begin{bmatrix}
        -S(^{\frameB}\omega_{\frameB}) {^{\frameMixed} {h}^{p}} + A_{lin}(s)T + mg{^{\frameB}R_{\frameW}e_{3}} \\
        -S(^{\frameB}\omega_{\frameB}) {^{\frameMixed} {h}^{w}} + A_{ang}(s)T
    \end{bmatrix}.
    \label{eq:centroidalMomBodyDerivative2}
\end{equation}
Expressing \eqref{eq:centroidalMom} in the $\frameMixed$ coordinates enables the matrices ${A}_{lin}({s})$ and ${A}_{ang}({s})$ to be represented exclusively as functions of the joints' internal configuration $s$. Specifically, the matrices depend on the rotation matrices between the base frame and each turbine frame, denoted as $^{\frameB}R_{J_i}, i \in {1,\dots,4}$, and the relative position vectors between the centre of mass and each turbine frame, $^{\frameB}p_{J_i, G}, i \in {1,\dots,4}$. Since these quantities are determined solely by the joints' configuration, this formulation facilitates the subsequent linearization of the matrices, as detailed in Section \ref{subsec:linearization}.\looseness=-1

\textbf{Assumption 1:} To simplify the equations of motion, we approximate the angular component of the centroidal momentum as:
\begin{equation}
    ^{\frameMixed} h^{w} \approx ^{\frameMixed} \mathbb{I} {^{\frameB}\omega_{\frameB}}, 
\end{equation} 
under the assumption that ${^{\frameB}\omega_{\frameCoM}} \approx {^{\frameB}\omega_{\frameB}}$. This approximation holds when the joint velocities are sufficiently small, a condition that is satisfied during flight, as the robot moves its arms slowly to adjust the jet directions \cite{nava2018flightController}.
With this assumption, the relationship between the derivative of the Euler angles $\dot{\phi} \in \mathbb{R}^{3}$ and the angular momentum $^{\frameMixed}h^{w}$ can be expressed as:
\begin{equation}
\dot{\phi} = (^{\frameMixed} \mathbb{I} W)^{-1} {^{\frameMixed}h^{w}},
\end{equation}
where $W$ denotes the transformation matrix mapping the body-frame angular velocity $^{\frameB}\omega_{\frameB}$ to the Euler angle~rates~$\dot{\phi}$~\cite{luukkonen2011modelling}.

\subsection{Dynamical model}

To model the jet dynamics, we adopt the nonlinear second-order model proposed in \cite{lerario2020modeling}, expressed as:
\begin{equation}
    \ddot{T}_{i} = h(T_{i},\dot{T}_{i}) + g(T_{i},\dot{T}_{i}) v(u_{th,i}),
    \label{eq:jetDynEquation}
\end{equation}
where $T_{i}$ is the thrust of the $i^{th}$ jet turbines, \( u_{th,i} \) represents the throttle of the $i^{th}$ jet turbines, and \( v(u_{th,i}) \) is a function designed to be invertible with respect to \( u_{th,i} \). For notational simplicity, we will refer to \( v(u_{th,i}) \) simply as \( v_{i} \).

By incorporating this jet model, the state vector \( z \) describing the dynamics of the robot is defined as:
\begin{equation}
    z = 
    \begin{bmatrix}
         {^{\frameW}x_{\frameCoM}} & {^{\frameMixed}h^{p}} & \phi & {^{\frameMixed}h^{w}} & T & \dot{T}
    \end{bmatrix}^{\top},
\end{equation}
where \( {^{\frameW}x_{\frameCoM}} \) denotes the centre of mass position, \( \phi \) represents the Euler angles, ${^{\frameMixed}h^{p}}$ and ${^{\frameMixed}h^{w}}$ are the linear and angular momentum described in Subsection \ref{subsec:centMomEq}, and \( T \) and \( \dot{T} \) denote the thrust and its rate of change, respectively.  
The control input vector \( u \) is given by:
\begin{equation}
    u = 
    \begin{bmatrix}
    s & v
    \end{bmatrix}^{\top},
\end{equation}
where \( s \) represents the joint positions, and \( v \) is the vector collecting the \( v_{i} \) auxiliary input related to the throttle command applied to the $i^{th} $ jet engines.

The system of equations that describe the dynamics of the mechanical system is as follows:
\begin{equation}
    \begin{cases}
        ^{\frameW}\dot{x}_{\frameCoM} = \frac{1}{m} {^{\frameW}R_{\frameB}} {^\frameMixed h^{p}} \\[3pt]
        ^\frameMixed \dot{h}^{p} = A_{lin}(s)T + mg{^{\frameB}R_{\frameW}}e_{3} - S(^{\frameB}\omega_{\frameB}) {{^{\frameMixed}h^{p}}} \\[3pt]
        \dot{\phi} =  {({^{\frameMixed} \mathbb{I}}  W)^{-1}} {{}^{\frameB} R_{\frameW}}  {^{\frameMixed} h^{w}}  \\[3pt]
        ^\frameMixed \dot{h}^{w} = {A}_{ang}({s}){T} - S(^{\frameB}\omega_{\frameB}) {^{\frameMixed} h^{w}} \\[3pt]
        \dot{T}_{i} = \dot{T}_{i} \\[3pt]
        \Ddot{T}_{i} = h(T_{i}, \dot{T}_{i}) + g(T_{i}, \dot{T}_{i}) {v_{i}} \\[3pt]
    \end{cases}
    \label{eqDyn}
\end{equation}
which we can rewrite in a compact form as:
\begin{equation}
    \dot{{z}} = f({z},{u}).
    \label{zDot}
\end{equation}
For the sake of clarity, in the following chapter, the superscript $\frameB$, $\frameW$ and $\frameMixed$ will be omitted.

\subsection{Linearised robot dynamics}
\label{subsec:linearization}
The resulting dynamic model of the robot is nonlinear with respect to the control inputs. However, solving a nonlinear MPC problem would significantly increase the computational burden and reduce the achievable control frequency. To address this issue, we choose to linearise the model, enabling the use of a linear MPC formulation and thus allowing for higher execution frequencies.

Specifically, we adopt a Linear Parameter-Varying (LPV) approach, where the dynamics are linearised around the current state \( z_{c} \) and control input \( u_{c} \). Using a first-order Taylor series expansion, the nonlinear dynamic equation \eqref{zDot} can be approximated as:
\begin{equation*}
    f(z,u) \approx f(z_c, u_c) + \left. \frac{\partial f}{\partial z} \right|_{z_c, u_c} \hspace{-0.9em}(z - z_c) + \left. \frac{\partial f}{\partial u} \right|_{z_c, u_c} \hspace{-0.9em}(u - u_c).
\end{equation*}

\textbf{Assumption 2:} In the linearization process, we assume that both the rotation matrix \( {^{\frameW}R_{\frameB}} \) and the skew-symmetric matrix \( S(\omega_{\frameB}) \) remain constant. This assumption is justified by the fact that, over the relatively short prediction horizon of the MPC, their variation is negligible. Additionally, the inertia matrix \( {^{\frameB} \mathbb{I}} \) is assumed constant, as it depends only on the joint positions, which are expected to change slowly within the horizon.

Under Assumptions 1 and 2, the linearised model is:
\begin{equation}
\begin{cases}
    \dot{x}_{\frameCoM} \approx \frac{1}{m} {^{\frameW}R_{\frameB}} h^{p} \\[3pt]
    \dot{h}^{p} \approx {A}_{lin}(s_{c}) T + m g {^{\frameB}R_{\frameW}} e_{3} + \lambda_{lin} (s - s_{c}) - S(\omega_{\frameB}) h^{p} \\[3pt]
    \dot{\phi} \approx \left({\mathbb{I}} W\right)^{-1} h^{w} \\[3pt]
    \dot{h}^{w} \approx {A}_{ang}(s_{c}) T + \lambda_{ang} (s - s_{c}) - S(\omega_{\frameB}) h^{w} \\[3pt]
    \dot{T}_{i} = \dot{T}_{i} \\[3pt]
    \begin{aligned}
        \ddot{T}_{i} \approx h(T_{c,i},\dot{T}_{c,i}) + \frac{\partial (h + g v)}{\partial T}\bigg|_{T_{c,i}, \dot{T}_{c,i}, v_{c,i}} (T_{i} - T_{c,i}) \\
        + \frac{\partial (h + g v)}{\partial \dot{T}}\bigg|_{T_{c,i}, \dot{T}_{c,i}, v_{c,i}} (\dot{T}_{i} - \dot{T}_{c,i}) + g(T_{c,i}, \dot{T}_{c,i}) v_{i}
    \end{aligned}
\end{cases}
\label{eq:DynLin}
\end{equation}
where the matrices \( \lambda_{lin} \) and \( \lambda_{ang} \) are defined as:
\begin{equation}
    \lambda_{lin} = \frac{\partial (A_{lin}(s) T)}{\partial s}\bigg|_{s_{c}, T_{c}}, \quad
    \lambda_{ang} = \frac{\partial (A_{ang}(s) T)}{\partial s}\bigg|_{s_{c}, T_{c}}.
    \label{eq:lambdaMatrices}
\end{equation}
Further details regarding the computation of these matrices are provided in Appendix \ref{appendix:linearization}.

The system \ref{eq:DynLin} can be rewritten as:
\begin{equation}
    \dot{z} \approx A(z,u) z + B(z,u)u + c(z,u).
    \label{eq:linearizedSys}
\end{equation}
where $c(z,u)$ represents the bias terms due to the linearisation. For the sake of clarity, in the following chapters we will refer to the matrices $A(z,u)$ and $B(z,u)$ and to the vector $c(z,u)$ evaluated in the current state of the robot as $A$, $B$ and $c$.

\subsection{MPC formulation}

In order to improve the tracking of a reference in position and orientation, the state $z$ is augmented to include two additional integral error variables $e_{x}$ and $e_{\phi}$ \cite{pannocchia2015offset}; the effect of adding those variables is to penalise the steady state offset. The augmented state thus becomes:

\begin{equation}
    z = 
    \begin{bmatrix}
         x_{\frameCoM} & h^{p} & \phi & h^{w} & T & \dot{T} & e_{x} & e_{\phi}
    \end{bmatrix}^{\top},
\end{equation}
whose dynamics is added to the system in \eqref{eq:DynLin} is defined as:
\begin{equation}
    \begin{cases}
        \dot{e}_{x} = (x_{\frameCoM} -x_{\frameCoM ,ref}) \\
        \dot{e}_{\phi} = (\phi -\phi_{ref})
    \end{cases}
\end{equation}
The following formulation for the linearised MPC is implemented:
\begin{align}
\begin{split}
   \min_{z, u} \quad & \| x_{\frameCoM} -  x_{\frameCoM, ref}\|^{2}_{W_{x}} + \| {h}^{p} -  {h}^{p}_{ref}\|^{2}_{W_{{h}^{p}}} \\
     + & \| \phi -  \phi_{ref}\|^{2}_{W_{\phi}}+ \| {h}^{w} -  {h}^{w}_{ref}\|^{2}_{W_{{h}^{w}}} \\
     + & \| \Delta u\|^{2}_{W_{u}} + \|e_{x} \|^{2}_{W_{e_{x}}} + \|e_{\phi} \|^{2}_{W_{e_{\phi}}} \\
    \text{subject to:} \quad \\
    z_{k+1} &{=} z_{k} {+} (A z_{k} {+} B u_{k} {+} c) \Delta t, \ \forall k = 1, \dots,N \\
    z_{0} &= z(t) \\
    s_{min} & \leq s_{k} \leq s_{max}, \ \forall k = 1, \dots,N \\
    v_{min} & \leq v_{k} \leq v_{max}, \ \forall k = 1, \dots,N
\end{split}
\label{eq:MPC}
\end{align}
Where $x_{\frameCoM, ref}$, ${h}^{p}_{ref}$, $\phi_{ref}$, and ${h}^{w}_{ref}$ are the reference of the centre of mass, linear momentum, Euler angles and angular momentum, and $W_{x}$, $W_{{h}^{p}}$, $W_{\phi}$, $W_{{h}^{w}}$, $W_{u}$, $W_{e_{x}}$ and $W_{e_{\phi}}$ are diagonal matrices that weight the tasks. The Forward Euler method is used to discretise and propagate the continuous dynamical system of the~\eqref{eq:linearizedSys}.\looseness=-1
\subsection{Multi-rate MPC}
\label{subsec:multirateMPC}

The considered jet-powered humanoid robot is equipped with two types of actuators: joints and jet engines. These actuators are controlled by low-level controllers operating at different frequencies. To account for the differences in actuator update rates, we modified the MPC formulation so that it generates new control inputs at frequencies that match those of the respective actuators.\looseness=-1

\begin{figure}
    \centering
    \resizebox{\linewidth}{!}{
        \input{pictures/multi_rate_propagation.tikz}
    }
    \caption{Multi-rate scheme; the red line represents the control input $u_{f}$, which is updated at every MPC step, while the blue line represents the control input $u_{s}$, which is updated every $N_{r}$ steps}
    \label{fig:multirate}
    \vspace{-0.32cm}
\end{figure}
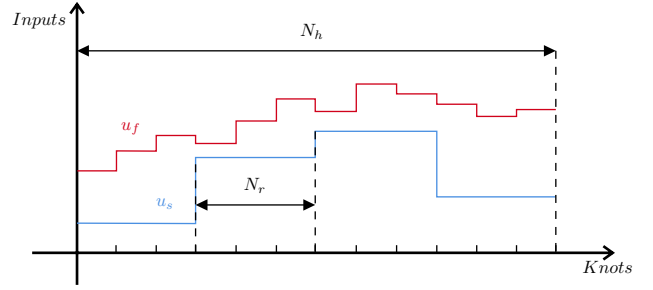

Let us assume that the MPC runs at a frequency \( f_{\text{MPC}} \) and that a given actuator operates at a lower frequency \( f_{s} \), with \( f_{\text{MPC}} > f_{s} \). Furthermore, suppose that \( f_{\text{MPC}} \) is an integer multiple of \( f_{s} \), such that \( N_{r} = \frac{f_{\text{MPC}}}{f_{s}} \). In this case, the number of control inputs \( N_{u_{s}} \) associated with the actuator within the prediction horizon is chosen as \( N_{u_{s}} = \frac{N_{h}}{N_{r}} \), where \(N_{h}\) is the horizon of the MPC, ensuring that the control input timing matches the actuator's update frequency. Figure \ref{fig:multirate} shows the variation of two different control inputs with two different control frequencies along the prediction horizon of the MPC. The red line represents a control input $u_{f}$ that is updated at every propagation step of the MPC, while the blue line represents a control input $u_{s}$ that is updated every $N_{r}$ steps of the MPC.\looseness=-1

Moreover, the control input \( {u}_{s} \) to be applied to the slow-frequency actuator is computed only at MPC iterations satisfying \( j = i \cdot N_{r}, \ i \in \mathbb{N} \). For all other MPC iterations, the first value \( u_{s,0} \) is constrained to be equal to the last control input computed for the actuator. Algorithm~\ref{alg:multi-rateMPC} outlines the multi-rate MPC approach developed to handle actuators operating at different update frequencies. The control inputs for slow actuators are recomputed only at specific MPC iterations aligned with their sampling periods, while previously computed inputs are maintained in the intermediate steps to ensure consistency.
\begin{algorithm}
\caption{Multi-Rate MPC}
\begin{algorithmic}[1]
\label{alg:multi-rateMPC}
\REQUIRE $f_{\text{MPC}}$, $f_s$, $N_r$, $u_{\text{prev}}, j$
\STATE $N_r \gets f_{\text{MPC}} / f_s$
\STATE $j \gets N_{r}$
\FOR{each MPC iteration $j$}
    \IF{$j = N_r$}
        \STATE Solve MPC problem
        \STATE Extract and apply $u_{s,0}$ for slow actuators
        \STATE $u_{\text{prev}} \gets u_{s,0}$
    \ELSE
        \STATE Constrain $u_{s,0} = u_{\text{prev}}$ in MPC
        \STATE Solve MPC problem
    \ENDIF
    \STATE Extract and apply remaining actuator commands
\ENDFOR
\end{algorithmic}
\end{algorithm}

%% file: pictures/multi_rate_propagation.tikz
\tikzset{every picture/.style={line width=0.75pt}} 

\begin{tikzpicture}[x=0.75pt,y=0.75pt,yscale=-1,xscale=1]

\draw [line width=1.5]  (127.42,221.16) -- (562.17,221.16)(161.02,34.77) -- (161.02,245.18) (555.17,216.16) -- (562.17,221.16) -- (555.17,226.16) (156.02,41.77) -- (161.02,34.77) -- (166.02,41.77)  ;
\draw [color={rgb, 255:red, 74; green, 144; blue, 226 }  ,draw opacity=1 ]   (161,198.93) -- (249.8,199.22) -- (249.8,149.73) -- (339.4,149.73) -- (339.4,130.08) -- (430.43,130.08) -- (430.43,179.22) -- (519.57,179.22) ;
\draw [color={rgb, 255:red, 208; green, 2; blue, 27 }  ,draw opacity=1 ]   (161.2,159.73) -- (190.56,159.73) -- (190.56,144.77) -- (220.33,144.93) -- (220.33,133.22) -- (249.8,133.22) -- (249.8,139.22) -- (280.14,139.22) -- (280.14,122.36) -- (310.33,122.36) -- (310.33,105.82) -- (339.4,105.82) -- (339.4,115.22) -- (370.14,115.22) -- (370.14,94.65) -- (400.33,94.65) -- (400.33,102.04) -- (430.43,102.04) -- (430.43,109.79) -- (460.33,109.79) -- (460.33,118.93) -- (490.14,118.93) -- (490.14,113.79) -- (519.57,113.79) ;
\draw    (190.22,215.43) -- (190.22,220.77) ;
\draw    (220.33,215.43) -- (220.33,221.43) ;
\draw    (249.8,215.43) -- (249.8,220.81) ;
\draw    (280.14,215.43) -- (280.14,221.31) ;
\draw    (310.33,215.43) -- (310.33,220.56) ;
\draw    (339.4,215.43) -- (339.4,221.06) ;
\draw    (370.14,215.43) -- (370.14,221.06) ;
\draw    (400.33,215.43) -- (400.33,221.31) ;
\draw    (430.43,215.43) -- (430.43,220.81) ;
\draw    (460.33,215.43) -- (460.33,221.56) ;
\draw    (490.14,215.43) -- (490.14,221.56) ;
\draw    (519.57,215.43) -- (519.57,220.81) ;
\draw  [dash pattern={on 4.5pt off 4.5pt}]  (249.8,220.81) -- (249.8,149.73) ;
\draw  [dash pattern={on 4.5pt off 4.5pt}]  (339.4,130.08) -- (339.4,220.33) ;
\draw    (252.8,185.27) -- (336.4,185.27) ;
\draw [shift={(339.4,185.27)}, rotate = 180] [fill={rgb, 255:red, 0; green, 0; blue, 0 }  ][line width=0.08]  [draw opacity=0] (8.93,-4.29) -- (0,0) -- (8.93,4.29) -- cycle    ;
\draw [shift={(249.8,185.27)}, rotate = 0] [fill={rgb, 255:red, 0; green, 0; blue, 0 }  ][line width=0.08]  [draw opacity=0] (8.93,-4.29) -- (0,0) -- (8.93,4.29) -- cycle    ;
\draw  [dash pattern={on 4.5pt off 4.5pt}]  (519.57,69.77) -- (519.57,221.31) ;
\draw    (164,69.77) -- (516.57,69.77) ;
\draw [shift={(519.57,69.77)}, rotate = 180] [fill={rgb, 255:red, 0; green, 0; blue, 0 }  ][line width=0.08]  [draw opacity=0] (8.93,-4.29) -- (0,0) -- (8.93,4.29) -- cycle    ;
\draw [shift={(161,69.77)}, rotate = 0] [fill={rgb, 255:red, 0; green, 0; blue, 0 }  ][line width=0.08]  [draw opacity=0] (8.93,-4.29) -- (0,0) -- (8.93,4.29) -- cycle    ;

\draw (218.67,178.73) node [anchor=north west][inner sep=0.75pt]    {$\textcolor[rgb]{0.29,0.56,0.89}{u_{s}}$};
\draw (192.67,121.73) node [anchor=north west][inner sep=0.75pt]    {$\textcolor[rgb]{0.82,0.01,0.11}{u_{f}}$};
\draw (284.33,163.67) node [anchor=north west][inner sep=0.75pt]    {$N_{r}$};
\draw (326.67,48.17) node [anchor=north west][inner sep=0.75pt]    {$N_{h}$};
\draw (538,225.71) node [anchor=north west][inner sep=0.75pt]    {$Knots$};
\draw (111,39.6) node [anchor=north west][inner sep=0.75pt]    {$Inputs$};

\end{tikzpicture}

%% file: sections/results.tex
\section{RESULTS}
\label{sec:results}

In this section, we present the simulation results of the proposed method. We chose to perform a recovery from an external disturbance to prove the robustness of the controller, the tracking of a minimum jerk trajectory, and two ablation studies to determine the impact of the jet dynamics and the multi-rate architecture.

\subsection{Simulation Environment}
\begin{figure*}
    \centering
    \includegraphics[width=1.0\linewidth]{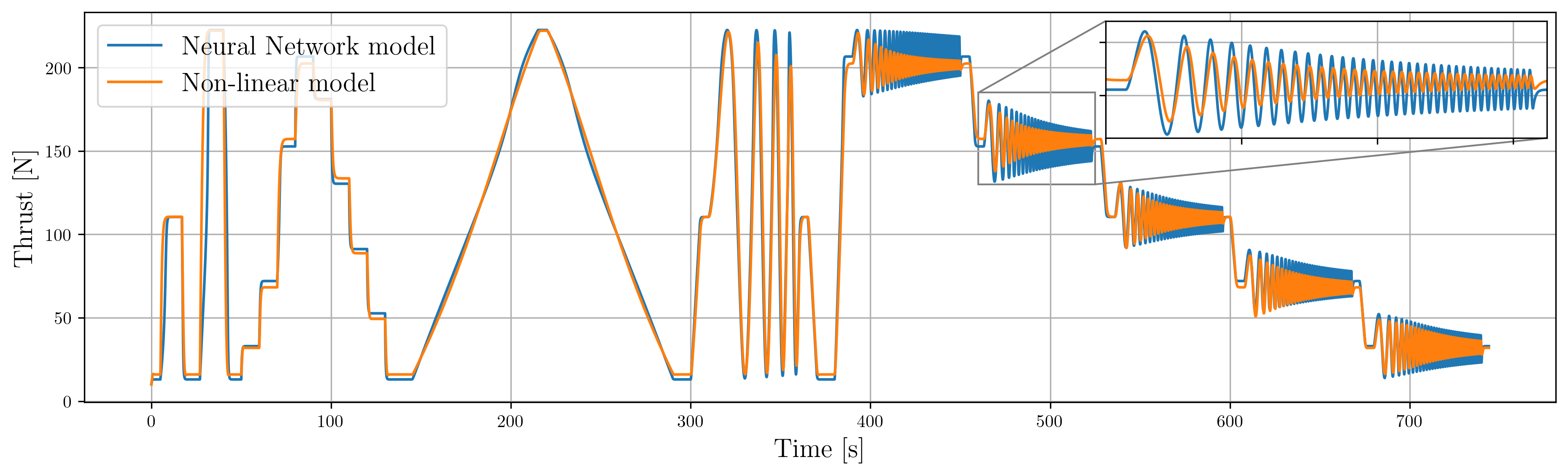}
    \caption{Neural Network and non-linear jet model comparison over the identification signal.}
    \label{fig:jet-non-linear-model}
\end{figure*}

The simulation environment used to perform the experiments is MuJoCo \cite{todorov2012mujoco}, running at a frequency of 1000~Hz. All simulations are conducted on an Ubuntu 22.04 operating system equipped with an Intel(R) Core(TM) i7-12700H processor\footnote{The code to run the simulations is available at \url{https://github.com/ami-iit/paper_gorbani_2025_humanoids_multi-rate-mpc-ironcub}}.
In order to have a simulation that is closer to reality, the jet dynamics are simulated using the neural network model proposed in \cite{lerario2024learning}, which is identified using data gathered from the real turbines. To identify the second-order model \eqref{eq:jetDynEquation}, the \textit{Extended Kalman Filter} (EKF) algorithm described in \cite{lerario2020modeling} is employed. Figure~\ref{fig:jet-non-linear-model} presents a comparison between the thrust predicted by the neural network model and the thrust generated by the identified non-linear model, when both are provided with the same throttle input profile. The mean absolute error (MAE) between the two models is approximately \( 5.95~\text{N} \). This comparison highlights the mismatch between the model used in simulation and one used in the prediction model of the MPC.\looseness=-1

In our case, the low-level controller of the joints operates at 1000~Hz, while the low-level controller of the jet engines operates at 10~Hz. Those values are chosen taking into consideration the specifics of the robot iRonCub. To enable the MPC to run at a higher frequency, we adopt the approach proposed in \cite{Lu2023VSMPC}, which adapts the sampling intervals of the MPC along the prediction horizon, with a shorter sampling time at the beginning of the horizon, which progressively increases towards the end. In fact, given the slow dynamics of the jet engines, we chose a control horizon of 1 second. Given that control horizon, to make the MPC run at 200 Hz with a constant timestep, we would need 200 knots, which would considerably slow down the problem. Exploiting this method, we can run the MPC at a frequency of 200~Hz with 17 knots. Recalling the notation in Subsection \ref{subsec:multirateMPC}, the low-frequency control $u_{s}$ input is the jet turbines control input $v$, which varies at a frequency of 10 Hz. The joint positions $s$ are computed at every step of the MPC, and the number $N_{r}$, since we adopt a variable sampling, varies along the control horizon, but it is chosen such that the jet turbine throttle is kept constant for \(0.1\) seconds.\looseness=-1

\subsection{Simulation results}

\subsubsection{Disturbance recovery}
\begin{figure*}
    \centering
    \includegraphics[width=1.0\linewidth]{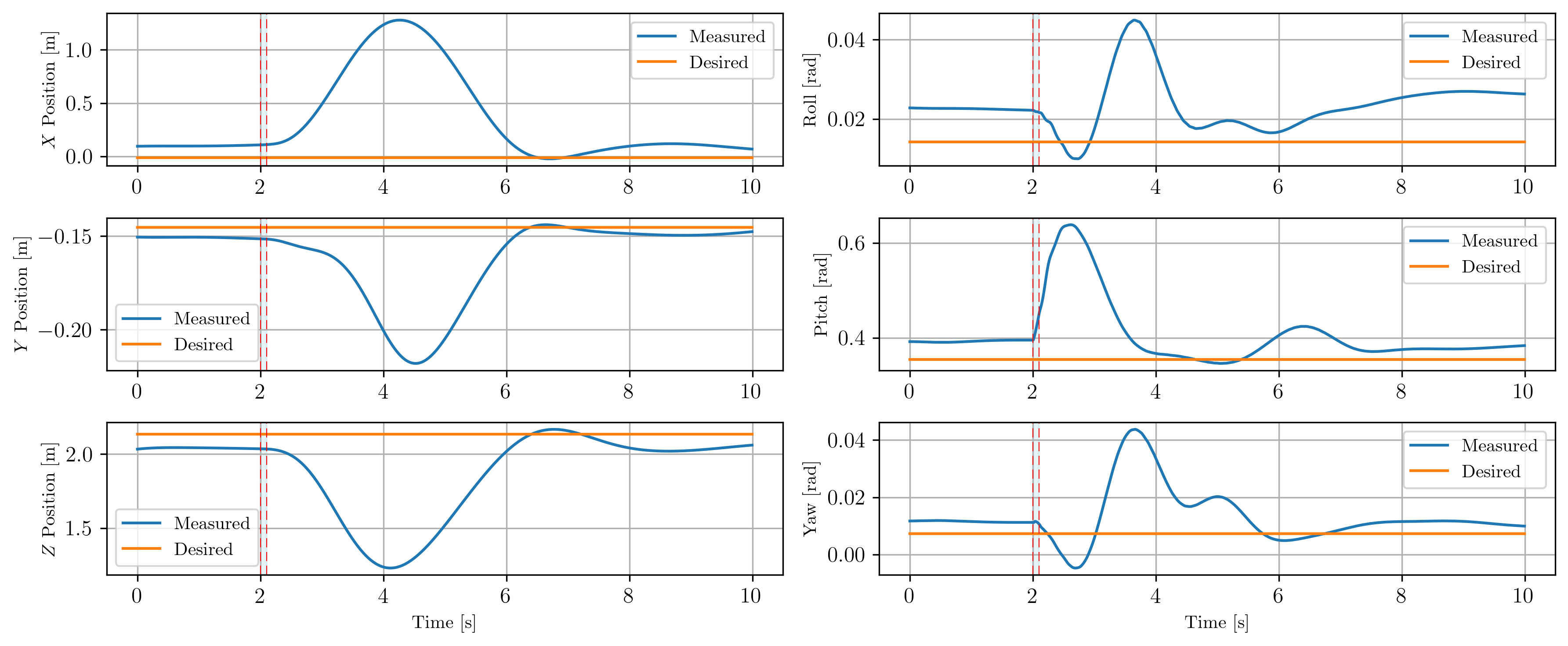}
    \caption{Centre of mass position and base orientation of the robot subject to an external disturbance applied between 2~s and 2.1~s (corresponding to the shaded region).}
    \label{fig:disturbance}
\end{figure*}

To demonstrate the controller's ability to reject external disturbances, an external torque of \( 300~\text{Nm} \) was applied along the \( y \)-axis, and an external force of \( 50~\text{N} \) was applied along the \( x \)-axis of the root link for \( 0.1~\text{seconds} \) at \( t = 2~\text{s} \). As shown in Figure~\ref{fig:disturbance}, the robot successfully recovers from the disturbance, despite experiencing a tilt of approximately \( 15^\circ \) in the roll axis, a displacement of about \( 1~\text{m} \) along the \( x \)-axis, and a vertical drop of approximately \( 1~\text{m} \).

\subsubsection{Minimum Jerk Trajectory Tracking}
\label{subsubsec:minJerk}
\begin{figure*}
    \centering
    \includegraphics[width=1.0\linewidth]{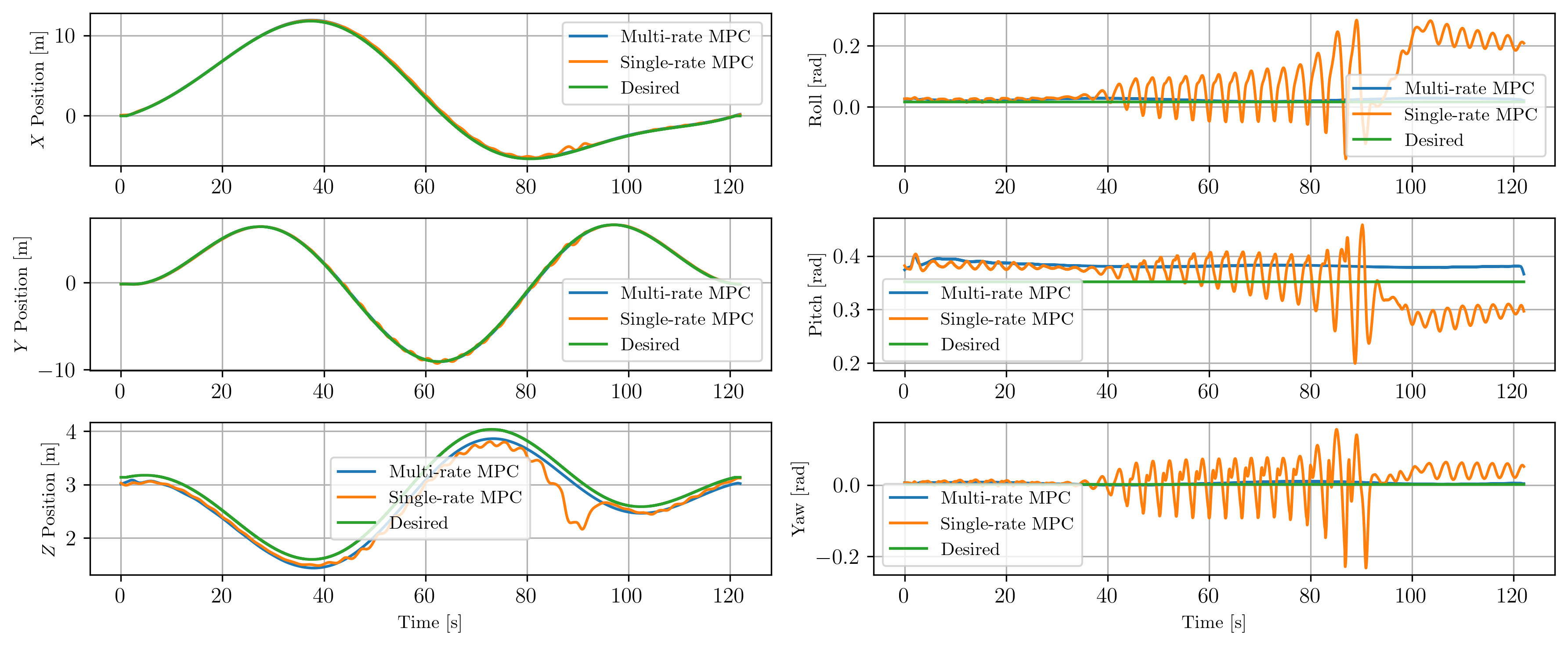}
    \caption{Comparison between the Multi-rate MPC and the Single-rate MPC in tracking a minimum jerk trajectory.}
    \label{fig:minimumJerk}
\end{figure*}

To test the performance of the MPC in tracking a reference trajectory, a minimum jerk trajectory is generated by interpolating a given set of points and used as the reference for the centre of mass position ${^{\frameW}x_{\frameCoM, ref}}$, while the reference orientation $\phi_{ref}$ is constant. The reference linear momentum is computed as $h^{l}_{ref} = m {^{\frameB}R_{\frameW}} {^{\frameW}\dot{x}_{\frameCoM, ref}}$, while the reference angular momentum $h^{w}_{ref}$ is set to zero since the reference orientation is constant. Figure \ref{fig:minimumJerk} shows the tracking of the centre of mass and the orientation of the base. The blue line is the measured signal, while the green one is the reference; the $x$ and $y$ components of the position are tracked fairly well by the controller, while the $z$ component presents an offset of about \( 0.15~\text{m} \). Concerning the orientation, on Roll and Yaw have a Mean Absolute Error lower than $0.01$~rad, while for the Pitch it is about $0.03$~rad. The error in the tracking is due to the discrepancies between the neural network model used to simulate the jet dynamics and the linearised non-linear model used in the prediction model of the MPC.

\subsection{Real-Time Feasibility}

\begin{figure}[h]
    \centering
    \includegraphics[width=1.0\linewidth]{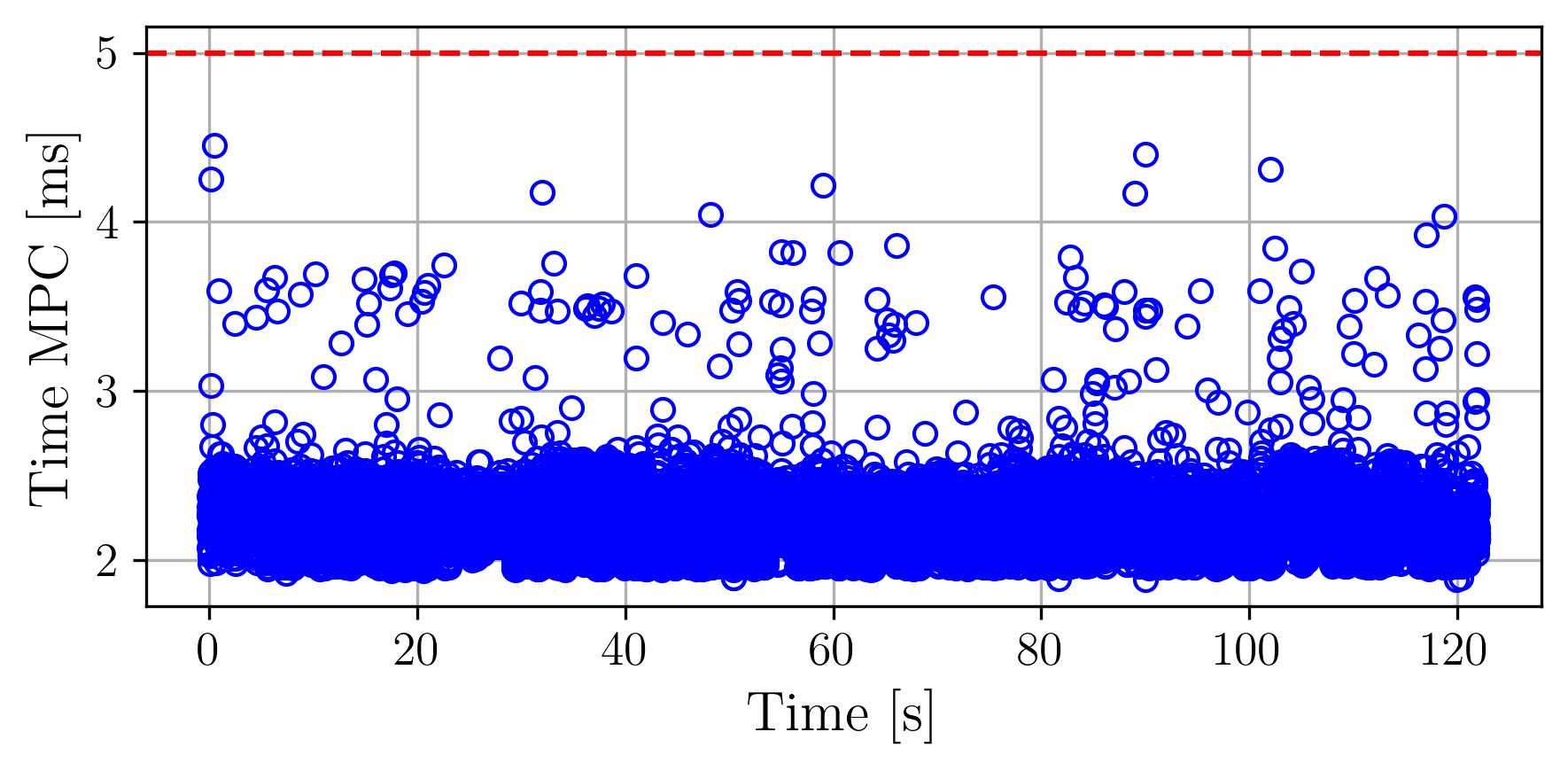}
    \caption{Time elapsed to solve each iteration of the MPC.}
    \label{fig:timeMPC}
    \vspace{-0.32cm}
\end{figure}

Ensuring real-time execution is critical for online MPC applications on flying robots. In our implementation, the MPC problem is formulated as a quadratic program (QP) and solved using the OSQP solver \cite{stellato2020osqp}. As shown in Figure~\ref{fig:timeMPC}, the average computation time per MPC iteration during tracking of the minimum jerk trajectory in Subsubsection \ref{subsubsec:minJerk} is 2.18~ms with a standard deviation of 0.16~ms and the maximum time of 4.45~ms, which is below the target frequency of 200~Hz (corresponding to a period of 5~ms). Given that the MPC is executed at 200~Hz (5~ms per cycle), the solver's performance is compatible with real-time operation. These results suggest that the proposed control strategy can feasibly run onboard for a complex system like iRonCub, provided that the onboard computational resources are comparable to those used in the simulation setup.\looseness=-1

\subsection{Ablation Study}
\subsubsection{Jet Dynamics study} To demonstrate the necessity of embedding the jet dynamics inside the MPC framework, we performed an ablation study where we removed the jet dynamics from the MPC and let the controller choose a thrust at will (subject to a regularisation cost); 
to compute the control input for the jet turbines, we used the Feedback Linearisation controller proposed in \cite{lerario2020modeling}. Testing the minimum jerk trajectory presented in \ref{subsubsec:minJerk}, the controller failed to stabilise the robot, and the robot fell on the ground.
\subsubsection{LPV study} To demonstrate the relevance of the LPV formulation, we carried out a simulation in which the model was linearised about the robot’s initial position and commanded to follow a minimum-jerk trajectory. In this case too, the robot was unable to track the trajectory and eventually fell.\looseness=-1

\subsubsection{Multi-rate study} To isolate the impact of the \textit{multi-rate} scheme, we compare our proposed controller with a \emph{single-rate} MPC variant. In the single-rate setup, the MPC computes jet thrust commands at every iteration (200~Hz), but only the commands at 10~Hz are applied, discarding the intermediate control inputs. This reflects the behaviour of the turbine's low-level controller, which operates at 10~Hz and ignores any commands received within each 0.1-second interval. Both controllers are evaluated on the minimum jerk trajectory described in Section~\ref{subsubsec:minJerk}. Figure~\ref{fig:minimumJerk} presents a comparison of the Multi-rate and Single-rate MPCs. While both achieve similar tracking performance in the $x$ and $y$ position components, the Single-rate MPC exhibits significantly greater oscillations in orientation. This observation is supported by the Mean Absolute Error (MAE) of the centre of mass position and base orientation reported in Table~\ref{tab:MAE}. These results highlight that explicitly accounting for actuator update rates in the MPC formulation leads to significantly improved tracking performance.\looseness=-1

\begin{table*}[]
\centering
\caption{Mean Absolute Error in tracking the minimum jerk trajectory for the multi-rate and single-rate MPC}
\label{tab:my-table}
\begin{tabular}{l|c|c|c|c|c|c}
\toprule
 & \textbf{X} & \textbf{Y} & \textbf{Z} & \textbf{Roll} & \textbf{Pitch} & \textbf{Yaw} \\
\midrule
\cellcolor{gray!10}Multi-rate  
  & \cellcolor{gray!10}0.1106 m & \cellcolor{gray!10}0.0729 m & \cellcolor{gray!10}0.1508 m 
  & \cellcolor{gray!10}0.0076 rad & \cellcolor{gray!10}0.0307 rad & \cellcolor{gray!10}0.0036 rad \\
 Single-rate 
  & 0.1429 m & 0.1005 m & 0.2034 m 
  & 0.0765 rad & 0.0371 rad & 0.0319 rad \\
\bottomrule
\end{tabular}
\label{tab:MAE}
\end{table*}

\subsection{Limitations}
One of the main limitations of the proposed MPC formulation is the reliance on Euler angles, which are inherently subject to singularities. However, given our choice of the reference frame in the case of the iRonCub robot, the singularity occurs only when the robot experiences a $90^\circ$ pitch rotation, i.e., when it becomes fully horizontal. Since our platform is designed for VTOL and avoids such horizontal flight, this theoretical limitation doesn't affect its intended operation.\looseness=-1

A second limitation arises from the linearisation process, as stated in Assumption 2, where the rotation matrix \( ^{\frameB}R_{\frameW} \) and the angular velocity $\omega_{\frameB}$ are assumed to be constant during the prediction horizon to maintain a linear problem structure. Given that the MPC horizon is set to 1 second and that the robot is not expected to perform aggressive manoeuvres within this time frame, it is reasonable to assume that the orientation and angular velocity remain sufficiently close to the initial condition, thereby justifying this simplification.

%% file: sections/conclusions.tex
\section{CONCLUSIONS}
\label{sec:conclusion}

In this paper, a momentum-based linearised Model Predictive Control (MPC) is proposed, which accounts for both the dynamics of the jet engines and the limitations in the control frequency of the jets. The controller demonstrates the ability to recover from external disturbances and track a smooth trajectory, even embedding in the MPC an approximated model of the jet dynamics with respect to the simulated one. Simulation-based validation is performed using the jet-powered humanoid robot iRonCub.

The proposed controller is a step forward in the realisation of a flight controller that works on the real robot, since it overcomes two of the limitations related to the hardware, which are the slow dynamics of the jet engines and the low frequency at which the turbines run.

However, several assumptions were made in the formulation of the controller. For example, the simplification of the robot's dynamics and the use of the Euler angles limit the applicability of this controller for aggressive manoeuvres. Additionally, aerodynamic forces are neglected in the current model.

Future work will involve testing the proposed controller on the real robot, as well as developing a specific controller capable of handling the take-off and landing phases.

%% file: sections/appendix.tex
\section*{APPENDIX}

\subsection{Linearization}

\label{appendix:linearization}

In this subsection, the computations needed to get the matrices in \eqref{eq:lambdaMatrices} are presented.

It is possible to compute the time derivative of the term $A_{lin}(s)T$ using the chain rule:
\begin{equation}
\frac{\partial A_{lin}(s)T}{\partial t} = \frac{\partial A_{lin}(s)T}{\partial s} \dot{s} + \frac{\partial A_{lin}(s)T}{\partial T} \dot{T}.
\label{eq:chainLambdaLin}
\end{equation}
The derivative can also be written by exploiting the definition of the derivative of a product:
\begin{equation}
    \frac{d A_{lin}(s)T}{d t} = \dot{A}_{lin}(s,\dot{s})T+A_{lin}(s)\dot{T}.
    \label{eq:derLambdaLin}
\end{equation}
Given that the \eqref{eq:chainLambdaLin} and \eqref{eq:derLambdaLin} are equal and since ${\frac{\partial A_{lin}(s)T}{\partial T} \dot{T} = A(s)\dot{T}}$, it holds that ${\frac{\partial A_{lin}(s)T}{\partial s} \dot{s} = \dot{A}(s,\dot{s})T}$. Assuming that the number of jets $n_{j}$ is 4, the matrix $\dot{A}(s,\dot{s})$ is ${}^{B}\dot{A} = \begin{bmatrix} {}^{B}\dot{R}_{J_1}e_3 & ... & {}^{B}\dot{R}_{J_4}e_3\end{bmatrix}$; exploiting the definition of the derivative of the rotation matrix we can rewrite the derivative as:
\begin{equation*}
    {}^{B}\dot{A} = \begin{bmatrix} S({}^{B}\omega_{B,J_{1}}){}^{B}{R}_{J_1}e_3 & ... & S({}^{B}\omega_{B,J_{n_j}}){}^{B}{R}_{J_{n_4}}e_3\end{bmatrix}.
\end{equation*}
Exploiting the property $S(x)y = -S(y)x$ and writing ${}^{B}\omega_{B,J_{j}} = J^{rel}_{\omega_j}(s)\dot{s}$ we can finally rewrite:
\begin{equation*}
{}^{B}\dot{A} = \begin{bmatrix} -S({}^{B}{R}_{J_1}e_3)J^{rel}_{\omega_1}\dot{s} & ... & -S({}^{B}{R}_{J_{n_4}}e_3)J^{rel}_{\omega_{n_j}}\dot{s} \end{bmatrix} ,    
\end{equation*}
and therefore it is possible to rearrange ${}^{B}\dot{A}(s,\dot{s})T$ as:
\begin{equation*}
    {}^{B}\dot{A}T = \begin{bmatrix} -T_1S({}^{B}{R}_{J_1}e_3) & ... & -T_{n_4}S({}^{B}{R}_{J_{n_4}}e_3) \end{bmatrix} \Bar{J}^{rel}_{\omega}\dot{s},
\end{equation*}
where $\Bar{J}^{rel}_{\omega}$ is a matrix composed by the Jacobians of the four jets defined as $\Bar{J}^{rel}_{\omega} = \begin{bmatrix} J^{rel, \top}_{\omega_1} & ... & J^{rel, \top}_{\omega_{4}} \end{bmatrix}^{\top} \in \mathbb{R}^{12 \times n_s}$, leading to the definition of $\lambda_{lin}(s,T)$ as:
\begin{equation*}
    \lambda(s,T) = \begin{bmatrix} -T_1S({}^{B}{R}_{J_1}e_3) & ... & -T_{n_j}S({}^{B}{R}_{J_{n_j}}e_3) \end{bmatrix} \Bar{J}^{rel}_{\omega}.
\end{equation*}
The matrix $\lambda_{ang}(s,T)$ can be found in an analogous way starting from the time derivative of the term $A_{ang}(s)T$, which by definition can be written as:
\begin{equation}
    \frac{d A_{lin}(s)T}{d t} = \dot{A}_{ang}(s,\dot{s})T+A_{ang}(s)\dot{T},
    \label{eq:derLambdaAng}
\end{equation}
exploiting the chain rule, the derivative can also be written as:
\begin{equation}
    \frac{d A_{ang}(s)T}{dt} = \frac{\partial A_{ang}(s)T}{\partial s} \dot{s} + \frac{\partial A_{ang}(s)T}{\partial T} \dot{T}.
    \label{eq:chainLambdaAng}
\end{equation}
Since \eqref{eq:derLambdaAng} and \eqref{eq:chainLambdaAng} are equal and $A_{ang}(s)\dot{T} = \frac{\partial A_{ang}(s)T}{\partial T} \dot{T}$, it holds that $\dot{A}_{ang}(s,\dot{s})T=\frac{\partial A_{ang}(s)T}{\partial s} \dot{s}$. The matrix $\dot{A}_{ang}(s,\dot{s})$ is:
\begin{equation}
    \dot{A}_{ang}(s,\dot{s}) = \begin{bmatrix} C_{1} & C_{2} & C_{3} & C_{4}  \end{bmatrix},
    \label{eq:AAngDerivative}
\end{equation}
where each column is defined as:
\begin{equation*}
    C_{i} = (S({}^{B}\dot{r}_{i}){}^{B}R_{i}e_3 + S({}^{B}r_{i}){}^{B}\dot{R}_{i}e_3), \ i \in \{1,2,3,4 \}.
\end{equation*}
where ${}^{B}r_{i}$ is the distance between the centre of mass and the $i^{th}$ turbine expressed in body frame. Exploiting once again the definition of the derivative of the rotation matrix, the property $S(x)y = -S(y)x$ and using the linear and angular Jacobians ${}^{B}\omega_{B, J_{j}} = J^{rel}_{\omega_j}(s)\dot{s}$ and ${}^{B}\omega_{B, J_{j}} = J^{rel}_{\omega_j}(s)\dot{s}$, \eqref{eq:AAngDerivative} rewrites as:
\begin{equation*}
    \dot{A}_{ang}(s,\dot{s}) =  - \begin{bmatrix} \bar{C}_{1} & \bar{C}_{2} & \bar{C}_{3} & \bar{C}_{4} \end{bmatrix},
\end{equation*}
where:
\begin{equation*}
    \bar{C}_{i} = (S({}^{B}R_{i}e_3){}^{B}J^{rel}_{\dot{r}_{i}} + S({}^{B}r_{i})S({}^{B}R_{i}e_3)J^{rel}_{\omega_{i}})\dot{s}.
\end{equation*}
The term $\dot{A}_{ang}(s,\dot{s})T$ can be rewritten isolating $\dot{s}$ as:
\begin{equation*}
\begin{aligned}
    &\dot{A}_{ang}(s,\dot{s}) T = \\
    &-(T_1(S({}^{B}R_{1}e_3){}^{B}J^{rel}_{\dot{r}_1} + S({}^{B}r_{1})S({}^{B}R_{1}e_3)J^{rel}_{\omega_1}) + ... \\ 
    & +T_{4}(S({}^{B}R_{4}e_3){}^{B}J^{rel}_{\dot{r}_{4}} + S({}^{B}r_{4})S({}^{B}R_{4}e_3)J^{rel}_{\omega_{4}}))\dot{s} \\
    &= \Lambda_{ang}(s,T) \dot{s}.
\end{aligned}
\end{equation*}